# AskBeacon – Performing genomic data exchange and analytics with natural language

Anuradha Wickramarachchi[1], Shakila Tonni[2], Sonali Majumdar[2], Sarvnaz Karimi[2], Sulev Kõks[3,4], Brendan Hosking[7], Jordi Rambla[5,6], Natalie A. Twine[7,8], Yatish Jain[7,8], Denis C. Bauer[1,8,9]*

[1] Australian e-Health Research Centre, Commonwealth Scientific and Industrial Research Organisation, Adelaide, Australia
[2] Data61, Commonwealth Scientific and Industrial Research Organisation, Sydney, Australia
[3] Centre for Molecular Medicine and Innovative Therapeutics, Murdoch University, Perth, WA 6150, Australia
[4] Perron Institute for Neurological and Translational Science, Perth, WA 6009, Australia
[5] Centre for Genomic Regulation (CRG), The Barcelona Institute of Science and Technology, Barcelona, Spain
[6] Department of Medicine and Life Sciences, Universitat Pompeu Fabra, PRBB, Barcelona, Spain
[7] Australian e-Health Research Centre, Commonwealth Scientific and Industrial Research Organisation, Sydney
[8] Macquarie University, Applied BioSciences, Faculty of Science and Engineering, Macquarie Park, Australia
[9] University of Sydney, School - School of Medical Sciences, Department of Biomedical Informatics and Digital Health, Sydney, Australia

**Enabling clinicians and researchers to directly interact with global genomic data resources by removing technological barriers is vital for medical genomics. AskBeacon enables Large Language Models to be applied to securely shared cohorts via the GA4GH Beacon protocol. By simply "asking" Beacon, actionable insights can be gained, analyzed and made publication-ready.**

GA4GH (Global Alliance for Genomics and Health) introduced the Beacon protocol[1] to standardise the exchange of genomic and phenotypic information. The underlying schema is designed for clinical and research use and enables industry-standard security and data governance practices. Beacon can jointly query genotypic and phenotypic data, with metadata information encoded through ontologies. However, empowering the community to perform these advanced queries requires a user-interface that can hide the underlying complexities, such as query combinations across collections (Cohorts and Datasets), entity types (Individuals, Biosamples, Runs, Analyses and Genomic variants) and return types (Records, Boolean, Counts), as well as translating medical or colloquial terminology to an ontology code, for example, SNOMED.

We developed AskBeacon to abstract the complexities of the Beacon schema using large language models (LLMs). AskBeacon is a web interface (**Supplementary Section 1**) on top of sBeacon[2], a cloud-based production-ready implementation of the Beacon protocol. It acts as the interpreter between a clinical or research question and the Beacon schema-formatted query (**Figure 1**).

For example, using AskBeacon, clinicians and researchers can validate in their data whether the genetically determined sex differences in Parkinson's Disease [3] are due to X-linked (*RPL10*[4]) or autosomal (*SNCA*[5]) genetic factors. Using natural language, users can perform currently expert-only steps such as, a) translating "Parkinson's disease" to an ontology code, b) identifying individuals in the cohort with the relevant genotypes c) constructing the right query to obtain the data, and d) analysing the data using custom scripts to visualize the results.

AskBeacon can query across federated repositories[6] in the global Beacon Network[7], where each node can be stood up securely and efficiently using an implementation of the Beacon protocol[8], like sBeacon. This enables even small clinical and research groups, e.g. from underrepresented populations, to share genomic data and enable the secure, fully consented, and controlled query across human genetic diversity (**Supplementary Section 2**).

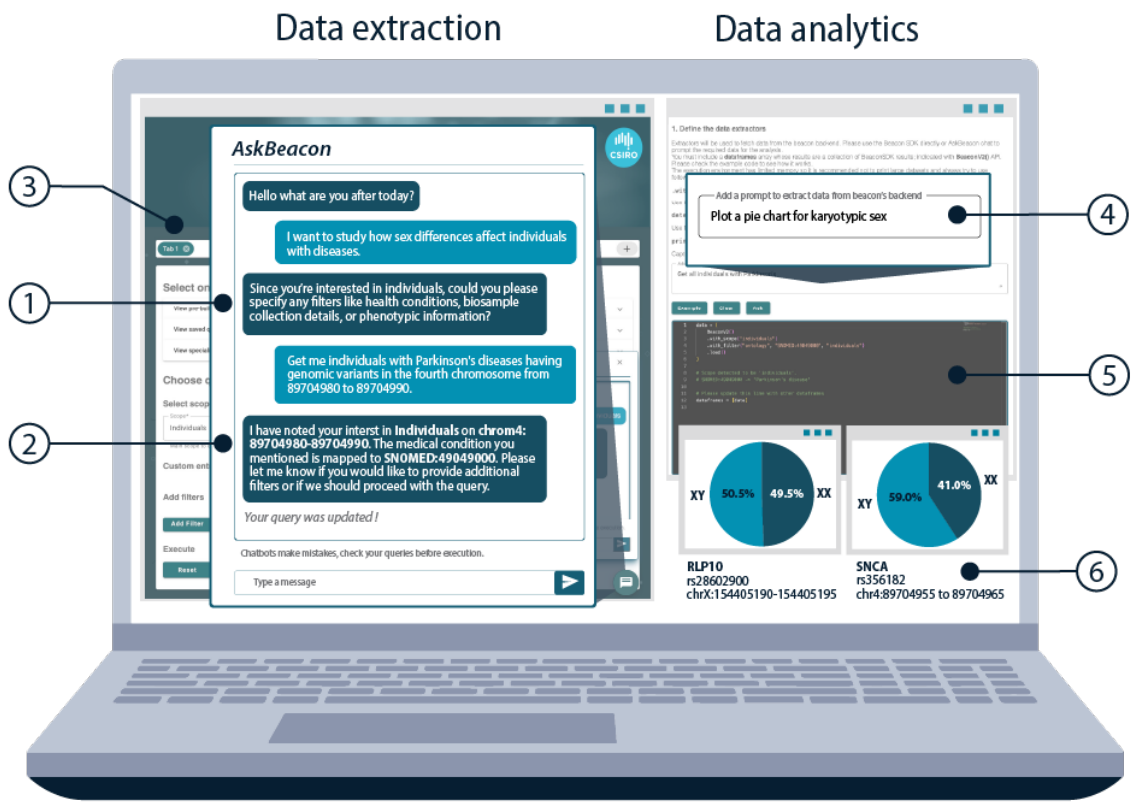

**Figure 1. Gaining actionable insights on genomic data**

For the research question "I want to study how sex differences affect individuals with disease", AskBeacon guides the user with prompts until all information for a successful Beacon query is presented (**Figure 1.1**). AskBeacon does the heavy lifting of obtaining ontology terms, interpreting genomic locations, inferring the query structure, and confirming all of this with the user (**Figure 1.2**). The actual query is executed by Beacon's software development kit (SDK, **Supplementary Section 6**), guaranteeing deterministic results. In our example, the data extraction obtains

Individuals with mutations in rs28602900 (*RPL10*) and rs356181 (*SNCA*) from the Parkinson's Progression Markers Initiative[9] (http://www.ppmi-info.org see acknowledgement). When the beacon instance provides a precise numeric answer, AskBeacon also supports the data analytics with prompts such as "Plot a pie chart for karyotypic sex" (**Figure 1.4**). It automatically generates the required analysis code and displays it for the user to confirm or adjust (**Figure 1.5**). In our case, **Figure 1.6** shows that the X-linked marker (*RPL10)* has no sex difference in PPMI, while the autosomal marker on chromosome 4 occurs 1.4 times more often in males with Parkinson's than females.

**Modular use of LLM to keep pace with innovation**

AskBeacon is methodology agnostic, enabling new LLMs to be added as they become available. Here, we tested the chat facility of currently available open (Ollama[10], HuggingFace[11]) and commercial or closed-weighted models (OpenAI models GPT3.5[12] and GPT 4[13], Anthropic model Claude 3.5[14]), as well as two different architectures (parallel and multi-step) for extracting the necessary information to populate a valid Beacon query (**Supplementary Section 3**). Specifically, we tested the individual components of a successful query, such as scope extraction, granularity extraction, variants extraction and filter extraction as well as query validity (**Supplementary Section 4**).

The parallel extraction approach was more resilient to network failures, incorrect inputs, partial inputs and malformed LLM, however it consumes more tokens compared to multi-step extractor. The multi-step approach can instead pick the suitable next chain depending on previous input or terminate chains early resulting in lower token consumption. However, failures at any point can terminate the query without extracting any information (**Supplementary Section 4**). Parallel workflows better scale to future throughputs with the elevated token consumption offset by an expected decrease in cost.

Overall, Gemma 2 was the most suitable open model for both parallel and multistep workflows (**Supplementary Table 4.1 and 4.2**) with average F1 scores of 0.92 and 0.81 for scope, granularity, variant, and filter extraction whereas GPT models were the most suitable commercial models with F1 scores of 0.91 and 0.81 respectively. Gemma 2's superior performance in parallel workflow may be due to its training on knowledge distillation rather than the next token prediction of GPT-4[15]. Commercial models overall had greater performance than open-weight models, probably due to the larger model size and up-to-date knowledge of bioinformatics and genome beacons. We have also presented LLM specific analyses of performance in **Supplementary Section 5**.

**Context management for building progressively complex interactions across multiple topics**

AskBeacon manages chat histories to enable users to build on previous queries. These history objects contain the summary of a conversation, specifically the variants, filters, chosen scope and granularity. This enables AskBeacon to request additional information to generate a Beacon compliant query. As shown in **Figure 1.3** the sBeacon User Interface (UI) allows the addition of multiple tabs so different queries can be maintained in parallel, enabling concurrent but independent querying.

**AskBeacon keeps the human in the loop**

AskBeacon keeps data extraction separate from data analysis to enable checkpointing from the human expert. All inferences are verified with the user to ensure elements such as ontology terms or genomic locations queried are aligned with the research question (**Figure 1.2**). Furthermore, the generated code for the data transformation and plotting is presented in an editor view to be amended by the users as needed (**Figure 1.5**). Following the execution of code, both standard output and error streams are made available to aid debugging and assist in code improvements (**Supplementary Section 6**).

**Security measures**

The analysis and exchange of sensitive data demands strict guard-rails to prevent misuse, leakage or exploitation by malicious actors. While all LLM vendors assure data privacy and security contractually, AskBeacon augments this by keeping sBeacon as the conduit between the data and the LLM, so that data is never exposed to the LLM directly. sBeacon ensures the safety of underlying data through state-of-the-art data security, with user management adhering to GA4GH protocol guidelines. sBeacon also ensures that AskBeacon's data extractors operate at the same level of access as the active user, hence no data can be extracted through AskBeacon that the user does not already have access to. AskBeacon's analytics is protected by using static code analysis and sandboxing, preventing accidental exposure of source code, files or runtime variables. We also perform static code analysis before any code execution and remove code with adversarial effect.

**Conclusion**

We built AskBeacon to leverage the GA4GH's Beacon protocol by enabling genomic data analysis using natural language through LLMs across the global Beacon network. In the future, we will extend the conversation capability to combine data extraction and analytics task, while retaining our extensive human checkpointing. This will enable users to ask comparative questions between cohorts, with AskBeacon automatically choosing the best statistical and visualization method. Given that Beacon is a discovery tool, differences in return values need to be considered, e.g. while some beacons return individual genotypes others might return summary statistics that are incompatible with each other. This also extends to differences in ontologies. We will leverage efforts from the terminology community to translate between different dictionaries and develop principles for beacons to communicate across different abstraction levels.

AskBeacon's LLM templates and chains are available at https://github.com/aehrc/AskBeacon.

**Acknowledgements**

Data used in the preparation of this article were obtained on [2024-09-30] from the Parkinson's Progression Markers Initiative (PPMI) database (https://www.ppmi-info.org/access-data-

specimens/download-data), RRID:SCR_006431. For up-to-date information on the study, visit http://www.ppmi-info.org.

PPMI – a public-private partnership – is funded by the Michael J. Fox Foundation for Parkinson's Research and funding partners, including 4D Pharma, Abbvie, AcureX, Allergan, Amathus Therapeutics, Aligning Science Across Parkinson's, AskBio, Avid Radiopharmaceuticals, BIAL, BioArctic, Biogen, Biohaven, BioLegend, BlueRock Therapeutics, Bristol-Myers Squibb, Calico Labs, Capsida Biotherapeutics, Celgene, Cerevel Therapeutics, Coave Therapeutics, DaCapo Brainscience, Denali, Edmond J. Safra Foundation, Eli Lilly, Gain Therapeutics, GE HealthCare, Genentech, GSK, Golub Capital, Handl Therapeutics, Insitro, Jazz Pharmaceuticals, Johnson & Johnson Innovative Medicine, Lundbeck, Merck, Meso Scale Discovery, Mission Therapeutics, Neurocrine Biosciences, Neuron23, Neuropore, Pfizer, Piramal, Prevail Therapeutics, Roche, Sanofi, Servier, Sun Pharma Advanced Research Company, Takeda, Teva, UCB, Vanqua Bio, Verily, Voyager Therapeutics, the Weston Family Foundation and Yumanity Therapeutics.

# Supplementary Material

## Table of Contents

# 1. AskBeacon chat functionality

AskBeacon is a web interface that facilitates convenient querying of genome Beacons and runs on top of sBeacon implementation. AskBeacon and sBeacon are both mobile between cloud providers and are currently deployed in the Amazon Web Services (AWS) using an AWS serverless stack. AskBeacon operates using AWS Lambda, Athena, and S3 storage facility.

## 1.1 Example user queries

| User Query | Scope | Granularity | Variant fields | Phenotypic filters |
|---|---|---|---|---|
| Which individuals have been diagnosed with hereditary cancers? | individuals | record | - | hereditary cancers – scope: individuals |
| What variants are found on chromosome 7 between 500k to 510k? | genomic variants | record | chrom – 7 start – 500k end – 510k | - |
| What sequence alterations have been found in the EGFR gene related to cancers? | genomic variants | record | - | EGFR gene – scope: individuals |

Table 1.1 Example user queries and relevant fields for extraction

## 1.2 Format of Beacon Payloads

For each filter, a type, a scope and a filter value can be specified. Supported filter types are "ontology", "alphanumeric" and "custom". The final request format is illustrated below.

```
{
  "query": {
    "filters": [],
    "requestedGranularity": "record",
    "requestParameters": {
      "assemblyId": "GRCh38",
      "start": [
        "110000"
      ],
      "end": [
        "110100"
      ],
      "referenceName": "1",
      "referenceBases": "N",
      "alternateBases": "N"
    }
  }
}
```

```
{
  "query": {
    "filters": [
      {
        "scope": "g_variants",
        "id": "SNOMED: 36340605",
        "value": "%colon cancer%"
      }
    ],
    "requestedGranularity": "record",
    "requestParameters": {
      "assemblyId": "GRCh38",
      "geneId": "%APC%"
    }
  }
}
```

## 1.3 Evaluation Dataset Details

We created four datasets that closely depicted questions that can be answered using a Beacon API. Dataset contained 150 scope extraction questions, 100 variant queries, 100 queries with filters and 100 granularity extraction questions. Apart from these questions, we also generated additional 50 irrelevant questions to evaluate the performance of our validator chain. The spreadsheet containing the complete set of questions is provided under the name "test_questions.xlsx". Each sheet contains questions, that would be the input from the user to AskBeacon and the parameters, the ideal pipeline should extract. The spreadsheet legend is provided below.

- Scope – questions to evaluate the performance of extracting the scope ('individuals', 'biosamples' or 'genomic variants').
- Variants – fields specific to the genomic variants labelled as the chromosome number, variation start position and variation end position
- Filtering terms – phenotypic filters that user intends to apply for the data extraction.
- Granularity – question to extract the level of details the user is asking for, the granularity labels can be either record, boolean or count.
- Invalids – a set of invalid/irrelevant question to the beacon context to measure the performance of rejecting out of scope questions by the user.

## 2. Access to data sources

Throughout the AskBeacon implementation, the data is accessed through the Beacon API to respond to user queries and for AskBeacon analytics. In AskBeacon implementation we use sBeacon as the Beacon protocol implementation. sBeacon has a secured API that requires users to be authenticated. AskBeacon uses the same authentication mechanism which enables users to interact with AskBeacon and sBeacon seamlessly. For AskBeacon analytics, the execution environment calls the sBeacon API behind the scenes using the same active session of the requesting user. This is achieved by reusing JWT tokens of the querying user to call sBeacon API. This ensures that the users can never use AskBeacon to reveal any information that they otherwise do not have access to.

# 3. Extractor implementation details

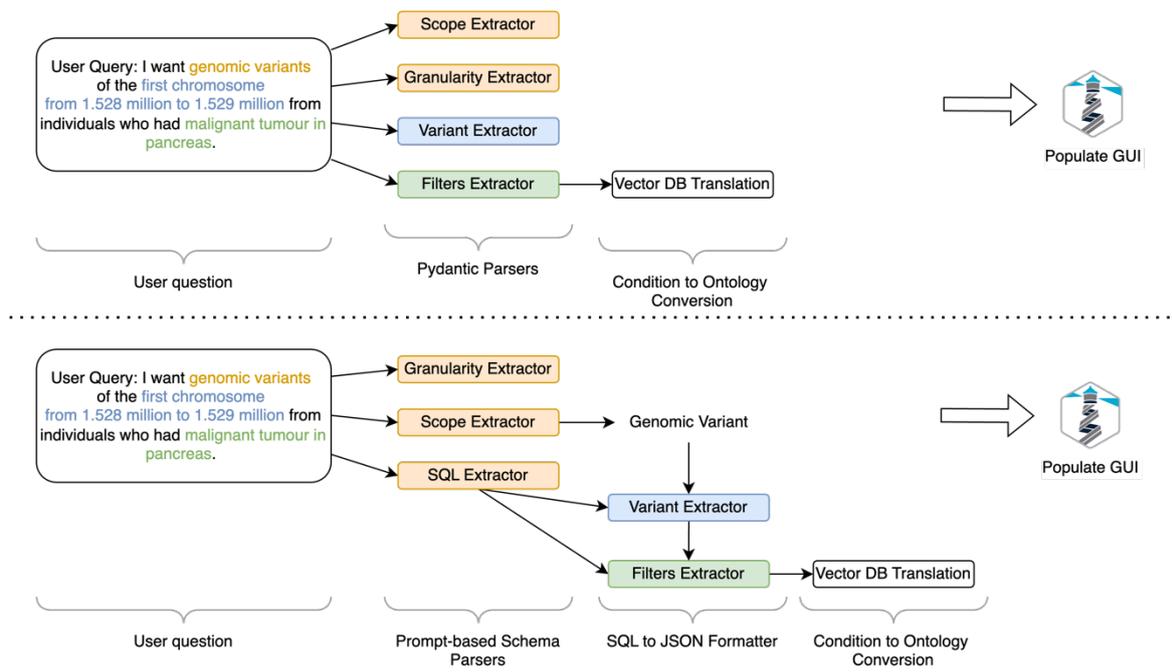

Figure 3.1 AskBeacon extractor chains architecture

## 3.1 Parallel workflow

We used one template for user question validation and four different extractor templates to extract scope, granularity, variant parameters and query filters.

**Scope extraction template**

```
INSTRUCTIONS
Given a user query, select the result scope to best suited for the query.
Return a JSON object formatted to look like:
{{
    scope: most appropriate scope to respond to the query, if unsure put "unknown"
}}

SCOPE MUST BE ONE OF
individuals: user expects individual or people entries to be returned
biosamples: user expects biosample or samples entries to be returned
runs: user explicitly mention runs entries to be returned
analyses: user explicitly mention analyses entries to be returned
datasets: user explicitly mention datasets entries to be returned
cohorts: user explicitly mention cohorts entries to be returned
g_variants: user explicitly expects only genomic variants or variants entries to be returned
unknown: if you cannot confidently pick any of the above

QUERY
{query}

OUTPUT
```

## Granularity extraction templates

```
INSTRUCTIONS
Does the query contain a set of conditions.
Return a JSON object with an array formatted to look like:
{{
    granularity: what kind of a return is expected, 'unknown' if not found
}}

CANDIDATE GRANULARITIES
record: requesting actual data records
count: asking for cardinality or count
boolean: only interested in the existence of data
unknown: if any of the above types not matched

QUERY
{query}

OUTPUT
```

## Variant parameter extraction template

```
INSTRUCTIONS
Given a user query, extract the assembly, chromosome, start base and end base.
Return a JSON object formatted to look like:
{{
    success: true or false, if a none of the attributes could be identified
    assembly_id: what is the mentioned assembly name (eg: grch38, hg38, or something similar)
                 if unsure use "unknown",
    chromosome: chromosome mentioned in the query (could be a numeric string from 1 to 22, X
                 or Y), if unsure use "unknown",
    start: start base position or interval; an array of one or two numbers,
                 if unsure use "unknown",
    end: end base position of interval; an array of one or two numbers,
                 if unsure use "unknown",
}}

CONDITIONS
whenever a field is unsure; use "unknown"
Ensure all fields are valid JSON

QUERY
{query}
```

**Filters extraction template**

```
INSTRUCTIONS
Does the query contain a set of conditions. Ignore anything that corresponds to a genomic variant
(position, chromosome, assembly)
Return a JSON object with an array formatted to look like:
{{
    filters: [
        {{
            term: place only one condition term here,
            scope: what might be the scope of this condition, chose from "individuals",
                "biosamples" and "runs"
        }}
    ]
}}

CONDITIONS
filters: this is an array with objects having two attributes called "term" and "scope".
        Ignore genomic variants
term: please insert only one condition here, if there are many, use multiple objects
        in "filters" array
if you do not see any terms, just return an empty array like {{filters: []}}

QUERY
{query}

OUTPUT
```

The above templates are formatted in Python using the {query} variable which places users' question in position. Note that double braces are used to pass through python string formatter as literal strings without injecting variables. The models were fed into Ollama and OpenAI python clients from Langchain library to evaluate open weight and commercial models respectively. All the API execution requests were made by setting the format flag to "json" to ensure structure output. The outputs were further validated using Pydantic Python library to ensure that AskBeacon throws exceptions upon failure. However, in out experiments we observed no such falsely structured JSON when the "json" flag was set with clear formatting instructions in the prompts.

**Validation Template**

```
You are a query builder for GA4GH Beacon V2. You must only accept user queries that are strictly
relevant to the GA4FH beacon V2 domain.
INSTRUCTIONS
Queries must be a biologically relevant question related to genomic variants, individuals,
biosamples, runs, analysis. Conditions can be anything related to health,
biosample collection or phenotypic information.

Consider the scenarios below.
- If user is just greeting, write a few words greeting along the lines "Hello, I am AskBeacon
assistant. What did you have mind for sBeacon?"
lightly modify the greeting message to be creative.

Respond with JSON of the form,
{{
    yes: true or false
    reason: reason behind yes/no - as shortly as possible. don't mention anything from user query.
}}

QUERY
{query}

OUTPUT
```

For validation, we used the above template, which returns a JSON structure that asserts the validity of the users' questions to the Beacon's scope. A reason for validity/invalidity is returned with the Boolean value to assist the generation of follow-up responses for the user.

## 3.2 Multistep Workflow

Below are the templates for scope and granularity extractions. For extracting the terms, we prompt the LLMs to generate the SQL statements aligning with the schema. Once the SQL statements are generated, we strip the filtered terms into genomic variants and phenotypic filter-related terms.

**Scope extraction template**

```
QUERY
{input}

INSTRUCTIONS
Classify the above user query into one of the below scopes:

individuals: The user expects individual or people entries to be returned
biosamples: The user expects biosample or sample entries to be returned
runs: The user explicitly mentions runs entries to be returned
analyses: The user explicitly mentions analysis entries to be returned
datasets: The user explicitly mentions datasets entries to be returned
cohorts: user explicitly mention cohorts entries to be returned
g_variants: user explicitly expects only genomic variants or variants entries to be returned
unknown: if you cannot confidently pick any of the above

The value of the scope should be 'individuals', 'biosamples', 'runs', 'analyses', 'datasets',
'cohorts' or 'g_variants'.
Your response should only contain the scope.

OUTPUT
"""
```

**Granularity extraction template**

```
QUERY
{input}

INSTRUCTIONS
Classify the above user query into one of the below categories:

record: requesting actual data records
count: asking for cardinality or count
boolean: only interested in the existence of data
unknown: if any of the above types are not matched

The value of the category should be 'record', 'count' or 'boolean'.
Your response should only contain the category.

OUTPUT
"""
```

## Text-to-SQL Template

```
{schema}

QUESTION
{input}

INSTRUCTIONS

Above question is regarding genomic variations.
Based on the schema create an SQL statement with "SELECT *" for the above question.
Any health condition, ethnicity, age, or gender are standard OntologyTerms.
If any field in the WHERE clause is an OntologyTerm, create a JOIN to OntologyTerm table.
Do not add additional information absent in the schema.
The response should contain only the SQL statement without any reasoning.

SQL QUERY:
"""
```

The schema used as part of the instruction is formatted as the BIRD text-to-SQL input data format[1]. BIRD (BIg Bench for LaRge-scale Database)[1] is a benchmarking dataset used in Text-to-SQL task evaluation. The schemas are prepared following beacon v2 schema definitions. The suitable schema is selected based on the scope extracted using the scope extraction template. A segment of the genomic variants schema formatted for the LLM instruction is illustrated below:

## Genomic variants schema

```
SCHEMA

GenomicVariations:
    variantInternalId:str [Example 'var00001', 'v110112']
    variation:LegacyVariation [REQUIRED] [Foreign key LegacyVariation.variantType]
    caseLevelData:CaseLevelVariant [Foreign key CaseLevelVariant.biosampleId]
    frequencyInPopulations:FrequencyInPopulations [Foreign key]
    FrequencyInPopulations.source]
    identifiers:str [Example one of clinvarVariantId, genomicHGVSId, proteinHGVSIds,
    transcriptHGVSIds or variantAlternativeIds (e.g. 'clinvar:12345', '9325')]
    molecularAttributes:MolecularAttributes [Foreign key
    MolecularAttributes.genomicFeatures or MolecularAttributes.geneIds]
    variantLevelData:PhenoClinicEffect [Foreign key PhenoClinicEffect.effect]

OntologyTerm: Ontology-based terms following CURIE syntax
    id:str [Example 'HP:0004789' or 'OMIM:164400']
    label:str [Example 'lactose intolerance','homozygous','hemizygous X-linked']

LegacyVariation : genomic variation information
    alternateBases:str [Format '^[ACGTUNRYSWKMBDHV\\-\\.]*$'] [DFLT 'N'] [REQUIRED]
    location:ChromosomeLocation [REQUIRED]
    referenceBases:str [Format '^[ACGTUNRYSWKMBDHV\\-\\.]*$'] [DFLT 'N'] [REQUIRED]
    variantType:str [DFLT 'SNP'] [REQUIRED]

ChromosomeLocation :
    species_id:str [DFLT taxonomy:9606]
    chr:str [One of 1,2,...,22,X,Y] [Chromosome number or name] [REQUIRED]
    start:int
    end:int
...
```

The resultant SQL statements are then run through SQLParse[2] library to extract the different field values and formatted as the beacon request. Following the parallel workflow, the API models - GPT-4-turbo[3] and GPT-3.5-turbo[4] - are also implemented using the OpenAI Python clients from the Langchain library. The open-source models - LLAMA3.1[5], GEMMA 2[6], FLANT5-Large, FLANT5-XL[7] and SQLCoder2[8] - are sourced from Huggingface[9] repository.

# 4. Task-specific Results

## 4.1 Parallel Extractors

The overall results obtained from the parallel extractors are illustrated in below table. Note that Precision, Recall and F1-Scores are computed using ROUGE[10] score values.

| Model | Metric | Scope extraction | Granularity extraction | Variants extraction | Filters extraction | Query validation |
|---|---|---|---|---|---|---|
| **Mistral Nemo**[11] | Precision | 0.903 | 1 | 0.928 | 0.727 | 0.833 |
|  | Recall | 0.897 | 1 | 0.958 | 0.926 | 0.833 |
|  | F1-score | 0.899 | **1** | **0.942** | 0.799 | 0.833 |
| **Llama 3.1** | Precision | 0.873 | 0.930 | 0.842 | 0.705 | 0.657 |
|  | Recall | 0.858 | 0.930 | 0.888 | 0.869 | 0.657 |
|  | F1-score | 0.864 | 0.930 | 0.863 | 0.755 | 0.657 |
| **Gemma 2** | Precision | 0.910 | 0.930 | 0.887 | 0.860 | 0.986 |
|  | Recall | 0.907 | 1 | 0.965 | 0.884 | 0.986 |
|  | F1-score | **0.908** | 1 | 0.922 | **0.855** | 0.986 |
| **Qwen 2**[12] | Precision | 0.867 | 1 | 0.857 | 0.778 | 0.934 |
|  | Recall | 0.857 | 0.875 | 0.940 | 0.790 | 0.934 |
|  | F1-score | 0.861 | 0.875 | 0.895 | 0.763 | 0.934 |
| **GPT-4 Turbo** | Precision | 0.873 | 0.875 | 0.968 | 0.752 | 0.998 |
|  | Recall | 0.872 | 1 | 0.968 | 0.913 | 0.998 |
|  | F1-score | 0.873 | **1** | **0.968** | **0.797** | 0.998 |
| **Claude 3.5 Sonnet** | Precision | 0.890 | 1 | 0.528 | 0.721 | 0.998 |
|  | Recall | 0.889 | 1 | 0.528 | 0.893 | 0.998 |
|  | F1-score | **0.889** | 1 | 0.528 | 0.779 | **0.998** |

Table 4.1: Precision, Recall and F1-scores of the models' predictions in parallel workflow. Best F1-score values are indicated in bold. Commercial models are shaded grey.

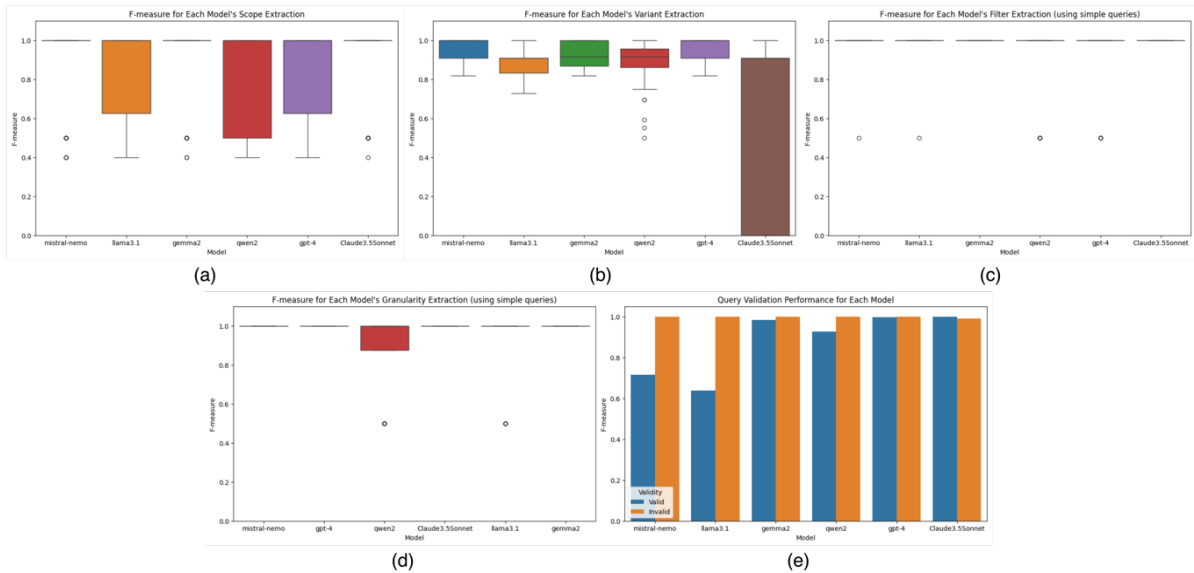

Figure 4.1: F1-scores comparison between parallel workflow models and their query validation performances (bottom-right)

The above plot illustrates the precision of detecting whether a question to AskBeacon is relevant or irrelevant, using open and closed-weighted models. Note that the results are presented under two criteria, using valid questions and invalid questions. As illustrated the questions where the context deviated from the Beacon capabilities were always detected to be invalid (precision 100%). However, the validity was falsely detected as invalid for nearly half of the questions using LLAMA3.1 and Mistral-Nemo. This is likely because these models were the smallest offerings of the kind, thus having little to no knowledge of the Beacon protocol. However, Gemma 2 and Qwen2 have more than 80% precision likely resulting from a better understanding of the prompt and reasonable knowledge of the Beacon protocol from their training data.

Furthermore, looking at the overall results in the table, we can compare the efficiency of using larger and smaller models. GPT 4 and Claude 3.5 can be considered the larger models used in parallel settings. However, interestingly, the other smaller models also perform close to these models. For scope extraction, with an F-score of 0.91, GEMMA 2 outperforms both GPT-4 (F-score 0.87) and CLAUDE3.5 (F-score 0.89) models. For the granularity extraction, Mistral Nemo and GEMMA 2 both models reach the F-score of 1 as GPT 4 and CLUADE 3.5 model.

For the complex tasks of extracting variants and filters, CLUADE 3.5 doesn't perform as well with only 0.53 and 0.78 F1-score for each of the tasks, respectively. GPT-4 is performing lower than GEMMA 2 in filter extractions. Overall, considering the size and utility trade-off, GEMMA 2 is the most suitable model for this workflow. The efficiency of GEMMA 2 may come from the training on knowledge distillation rather than the next token prediction of GPT-4.

## 4.2 Multistep Extractors

The overall results obtained from the multistep extractors are illustrated in below table.

| Model | Metric | Scope extraction | Granularity extraction | Variants extraction | Filters extraction |
|---|---|---|---|---|---|
| **FLANT5- Large** | Precision | 0.880 | 0.835 | n/a | n/a |
| | Recall | 0.880 | 0.835 | n/a | n/a |
| | F1-score | 0.880 | 0.835 | n/a | n/a |
| **FLANT5 - XL** | Precision | 0.940 | 0.875 | n/a | n/a |
| | Recall | 0.940 | 0.875 | n/a | n/a |
| | F1-score | **0.940** | 0.875 | n/a | n/a |
| **Gemma 2** | Precision | 0.933 | 0.926 | 1 | 0.779 |
| | Recall | 0.933 | 0.926 | 0.684 | 0.500 |
| | F1-score | 0.933 | **0.926** | 0.763 | **0.634** |
| **LLAMA 3.1** | Precision | 0.827 | 0.818 | 0.640 | 0.478 |
| | Recall | 0.827 | 0.818 | 0.529 | 0.378 |
| | F1-score | 0.827 | 0.818 | 0.557 | 0.400 |
| **SQLCoder 2** | Precision | n/a | n/a | 0.929 | 0.563 |
| | Recall | n/a | n/a | 0.838 | 0.452 |
| | F1-score | n/a | n/a | **0.861** | 0.508 |
| **GPT-3.5 Turbo** | Precision | 0.720 | 0.806 | 0.993 | 0.785 |
| | Recall | 0.720 | 0.806 | 0.984 | 0.656 |
| | F1-score | **0.720** | 0.806 | **0.988** | 0.721 |
| **GPT-4 Turbo** | Precision | 0.707 | 0.820 | 1 | 0.791 |
| | Recall | 0.707 | 0.820 | 0.956 | 0.479 |
| | F1-score | 0.707 | **0.820** | 0.978 | **0.735** |

Table 4.2: Precision, Recall and F1-scores of the models' predictions in multistep workflow. Best F1-score values are indicated in bold. Commercial models are shaded grey.

In the multistep workflow, classifying the user query into scope and granularity levels is critical for selecting the appropriate beacon schema for natural language to SQL query mapping. Like the parallel workflow, we observe that model efficiency improves relative to parameter size. For scope and granularity classification, the FLANT5-XL model (with just 3B parameters) performs best, achieving F1-scores of 0.94 and 0.88, respectively, outperforming larger models.

Unlike scope and granularity prediction, generating an SQL statement from a natural language query is more complex. This involves three-folded analysis by the LLMs – mapping text keywords to the schema, identifying keyword-parameter matches, and producing the SQL statement. From the generated SQL, we extract variant fields and filters. While this method produces more schema-aligned keywords and reduces hallucinated keywords, it is more challenging than generating straightforward text responses.

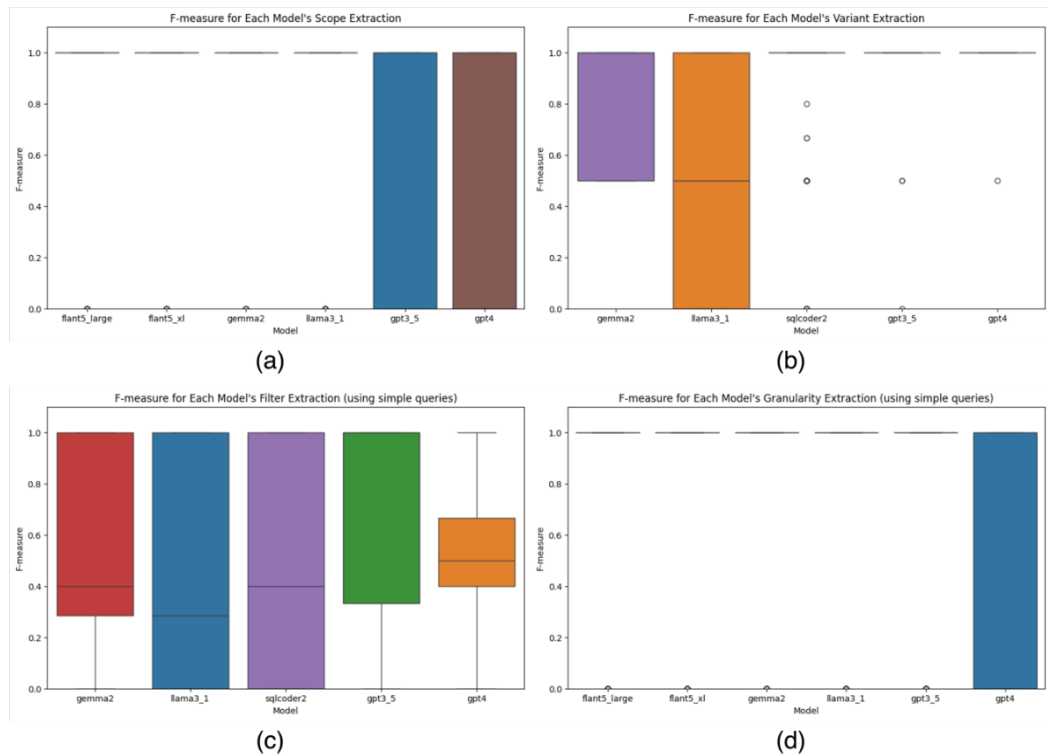

Figure 4.2: F-scores comparison between multistep workflow models

As shown in the table, GPT-3.5 performs best in extracting variants and phenotypic terms, while smaller models like LLAMA 3.1 and GEMMA 2 struggle with proper schema mapping. However, the SQLCoder 2 model, despite being smaller, outperforms LLAMA 3.1 and GEMMA 2 in variant extraction, likely due to fine-tuning on standard question-answering datasets. This highlights the merit of fine-tuning smaller models on domain-specific dataset, enabling them to generalize well to human genomic queries and potentially rival commercial models.

## 4.3 Parallel vs Multistep workflow

Comparing the parallel and multistep workflows, we can see that for the scope and granularity extraction, GEMMA 2 and FLANT5-XL LLMs are the best-performing models as depicted in the below table:

| Workflow | Scope extraction | Granularity extraction | Variants extraction | Filter extraction |
|---|---|---|---|---|
| **Parallel** | Gemma 2 F1-score: 0.91 | Gemma 2 F1-score: 1.00 | GPT-4 Turbo F1-score: 0.96 | Gemma 2 F1-score: 0.85 |
| **Multistep** | FLANT5-XL F1-score: 0.94 | Gemma 2 F1-score: 0.93 | GPT-3.5 Turbo F1-score: 0.98 | GPT-4 Turbo F1-score: 0.73 |

Table 4.3: Task-specific top performing models with their F1-scores

However, extracting the variants and filtering terms is the most important task in the overall workflow. We found, GEMMA 2 is the overall high performing model in parallel workflow with the highest F-score of 0.85 in filtering terms extraction and second high F-score of 0.92 in variant related terms extraction (Table 4.1). On the other hand, GPT-4 is the overall better performing model in multistep workflow with the top F-score of 0.73 in filtering terms extraction and second high F-score of 0.97 in extracting variants related terms (Table 4.2). Further discussion of the performance of the two workflows is given in the following sections.

## 5. LLM Performance Analysis

The performance analysis of the language models involves learning further in what ways the LLMs struggles to determine the fields.

### 5.1 Scope and Granularity Extraction

In determining the scopes and granularities of the user queries, the instruction specifically mentions to choose "unknown" if cannot be determined. The overall predictions of the "unknown" labels are extremely low and there are slightly more unknown labels predicted in determining the scopes than the granularity. Below table depicts the percentage of the predictions as unknown by different models.

| Workflow | Model | Scope extraction % | Granularity extraction % |
|---|---|---|---|
| Parallel | Claude 3 .5 Sonnet | 19.00 | 0.00 |
| | Gemma 2 Instruct | 9.00 | 0.00 |
| | LLAMA 3.1 Instruct | 0.00 | 0.00 |
| | GPT 4 Turbo | 22.00 | 0.00 |
| | Mistral Nemo | 1.00 | 0.00 |
| | QWEN 2 Instruct | 14.00 | 16.67 |
| Multistep | GPT 4.0 Turbo | 28.00 | 7.00 |
| | GPT 3.5 Turbo | 24.00 | 2.00 |
| | LLAMA 3.1 Instruct | 4.00 | 0.00 |
| | Gemma 2 Instruct | 4.00 | 0.00 |
| | FLANT5-Large | 4.00 | 0.00 |
| | FLANT5-XL | 1.00 | 0.00 |

Table 5.1 Rate of predicting "unknown" labels by the LLMs

We also further analyse scopes of which user queries are easily identifiable by the models. Below diagram shows the overall distribution of the original scope labels – individuals, biosamples and genomic variants – along with the percentage of each labels predicted accurately by the models. In both the workflows, we observe that the models are comparatively recognising genomic variants related queries with higher accuracy than the queries related to individuals or their biosamples.

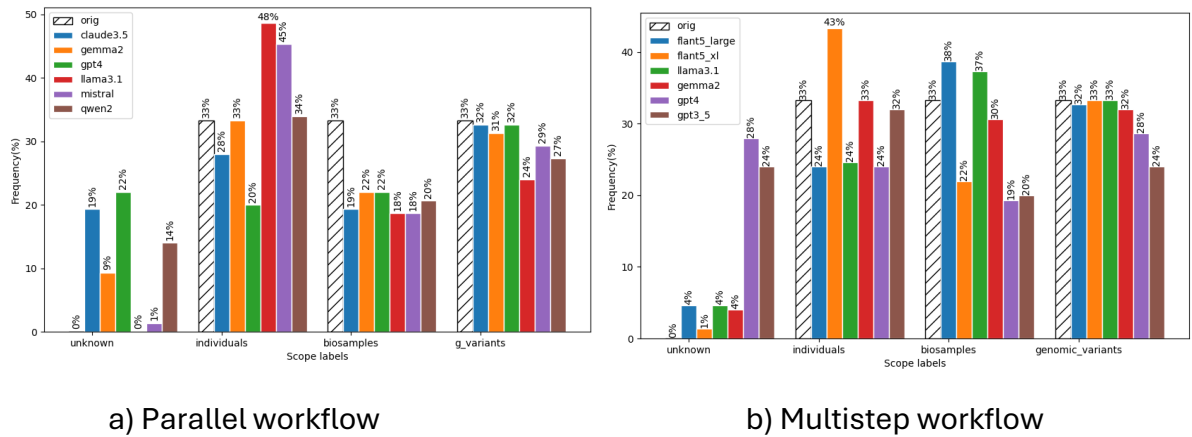

a) Parallel workflow                              b) Multistep workflow

Figure 5.1 LLMs in extracting the scopes of the user queries

## 5.2 Variant Terms Extraction

For the genomic variants related and other phenotypic term extractions, we report the rate of correctly extracted terms as the extraction accuracy and the rate of missed terms as the incompleteness. To better understand which model is working better at predicting what type of terms, we take the predictions of the different models and restrict the predictions by considering the subset of the user queries that has all three variant fields (variants complete) obtaining 71 user queries.

Below table shows the results. In the parallel workflow, Claude 3.5 model is missed 48.56% of the all the variant terms. In multistep workflow, we observe that two of the open-source models – LLAMA3.1 and GEMMA 2 – are failing to capture a significant percentage of the terms. This scenario is different from the scope and granularity extraction, where the open-source models are the better-performing ones. The incompleteness rate is slightly lower in the case of the SQLCoder2 model, as this model is further finetuned on the natural language to SQL question answering task.

| Workflow | Model | Extraction accuracy % | Incompleteness % |
|---|---|---|---|
| Parallel | Claude 3.5 Sonnet | 51.44 | 48.56 |
|  | Gemma 2 Instruct | 82.72 | 0.82 |
|  | LLAMA 3.1 Instruct | 99.18 | 0.82 |
|  | GPT 4 Turbo | 86.01 | 0.82 |
|  | Mistral Nemo | 92.18 | 0.82 |
|  | QWEN 2 Instruct | 73.66 | 2.88 |
| Multistep | GPT 4.0 Turbo | 99.32 | 0.68 |
|  | GPT-3.5 Turbo | 98.30 | 1.70 |
|  | LLAMA 3.1 Instruct | 41.50 | 58.50 |
| | Gemma 2 Instruct | 51.70 | 48.30 |
| | SqlCoder2 | 79.25 | 20.75 |

Table 5.2 Accuracy of genomic variant attribute extraction

We also further produce below figure that show the rate of extracting the chromosome number, variation start and end position. In the parallel workflow, we can see that in general models are struggling to predict the third variants field, end position, even though they predict the start position correctly more often. From the multistep workflow we can see that shows GPT-4 can predict all three genomic variants related fields, whereas GEMMA 2 can relate mostly to the chromosome fields to the schema, not their positions.

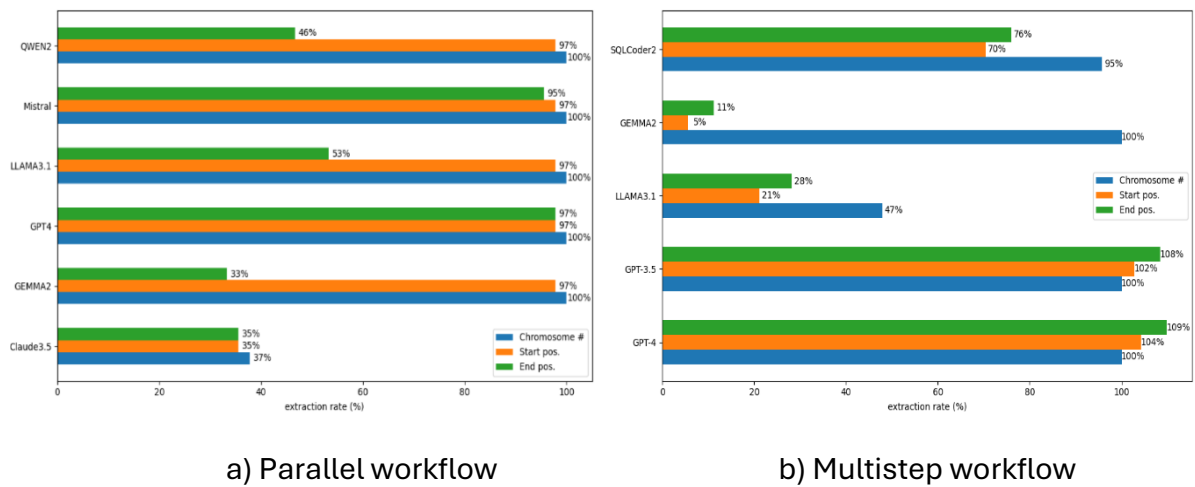

a) Parallel workflow      b) Multistep workflow

Figure 5.2 Extraction rate (%) of genomic variant attributes

## 5.3 Phenotypic Terms Extraction

For phenotypic term extractions as well, we consider only the user queries that have terms other than genomic variants related ones and report the rate of correctly extracted terms as the extraction accuracy and the rate of missed terms as the incompleteness. Below table shows the results for both parallel and multistep workflow. We observe more incompleteness in the multistep workflow than the parallel one phenotypic terms extraction.

| Workflow | Model | Extraction accuracy % | Incompleteness % |
|---|---|---|---|
| **Parallel** | Claude 3 .5 Sonnet | 73.00 | 5.00 |
| | Gemma 2 Instruct | 64.00 | 0.00 |
| | LLAMA 3.1 Instruct | 37.00 | 0.00 |
| | GPT 4 Turbo | **76.00** | 2.00 |

|  | Mistral Nemo | 48.00 | 0.00 |
|---|---|---|---|
|  | QWEN 2 Instruct | 34.00 | 0.00 |
| **Multistep** | GPT 4.0 Turbo | **80.58** | 19.42 |
|  | GPT-3.5 Turbo | 73.38 | 26.62 |
|  | LLAMA 3.1 Instruct | 34.89 | **65.11** |
|  | Gemma 2 Instruct | 53.96 | 46.04 |
|  | SqlCoder2 | 43.53 | 56.47 |

Table 5.3 Accuracy and incompleteness of phenotypic filter extraction

## 5.4 LLM Hallucination in Terms Extraction

Further evaluation of in which cases the models are introducing additional terms, as a measure of the LLM's hallucination check, we find that for variants terms extraction the rate of adding unknown terms is insignificant. On the other hand, for phenotypic terms extraction, only the LLAMA3.1 model is hallucinating in 17.9% of cases. However, the impact of missing terms and additional term introduction is handled by the beacon UI validations.

| Workflow | Model | Variants Terms Extraction % | Filtering Terms Extraction % |
|---|---|---|---|
| Parallel | Claude3 .5 | 0.00 | 22.00 |
|  | Gemma 2 Instruct | 16.46 | 36.00 |
|  | LLAMA 3.1 Instruct | 13.17 | 63.00 |
|  | GPT 4 Turbo | 0.00 | 22.00 |
|  | Mistral Nemo | 7.00 | 52.00 |
|  | Qwen 2 Instruct | **23.46** | **66.00** |
| Multistep | GPT 4.0 Turbo | 0.00 | 2.52 |
|  | GPT 3.5 Turbo | 0.00 | 5.04 |
|  | LLAMA 3.1 Instruct | 0.34 | **17.9** |
|  | Gemma 2 Instruct | 0.00 | 2.16 |
|  | SQLCoder 2 | 1.36 | 3.60 |

Table 5.4 LLMs hallucination analysis – unknown terms introduction (%)

## 5.5 Parallel and Multistep Workflow Comparison

For the two investigated workflows, there are significant difference in the prediction of variants terms and additional phenotypic filtering terms.

An intuitive comparison between the parallel and multistep extraction model is that, in the parallel workflow the models' instructions are rather simple, where the model is asked to predict only variants specific fields (variants extractor template) and other fields (filter extractor template) not concerning about the presence of the fields in the Beacon schema. Not all extracted terms in this extractor chain are valid for Beacon. A further validator template is further required here to filter out the terms that are not related to Beacon.

In contrast, in the multistep workflow, both the variants and phenotypic terms are extracted only when they match with the beacon schema without the necessity of the validation prompt. Thus, although these models are predicting less terms, the extracted terms are aligned with the schema with less hallucination than the Parallel schema, as seen in previous section.

# 6. AskBeacon analytics functionality

AskBeacon analysis has two steps 1) data extraction and 2) data analysis. We use BeaconSDK script to perform data extraction. The data analysis is performed using execution of another script on top of extracted data. Since both steps relies on code execution, we provide users with the standard output and standard errors to further facilitate debugging on their code. These outputs are captured using the native python context manager to redirect these output streams to be presented to the user.

## 6.1 Beacon SDK

Beacon SDK is a fluent API interface that wraps the complexity of Beacon queries for easy static code generation after the extraction of necessary fields using AskBeacon extractor chains. The interface is as follows.

```
data = (
    BeaconV2()  → initialise the SDK instance
    .with_scope(SCOPE)  → sets the scope for data extraction
    .with_filter(TYPE, FILTER, SCOPE)  → Attach filters
)

data  → this is a Pandas data frame
```

## 6.2 Analytics data extraction and execution

The SDK construction was driven by the same extractors we used in the AskBeacon chatbot. Upon extraction of data in JSON format from the Beacon API, it is transformed into a Pandas data frame with columns corresponding to each JSON attribute. The code generation for the analysis takes place using the following prompt.

```
INSTRUCTIONS
You must create a python code achieve the task indicated by the user query.
Please comment your code for easy understanding
The response must be in the following JSON format
{
    code: python code
    files: [] # this is a list of output files paths written by the script
    assumptions: [] # list of short assumptions you made
    feedback: [] # list of short instructions for user if they need to attend such as modify
                attributes according to the actual table content (do not include anything related
                to libraries, imports or variables)
}

CONDITIONS
You can only use following python libraries and they are already imported as shown within brackets
* Pandas (import pandas as pd)
* Numpy (import numpy as np)
* Matplotlib (import matplotlib.pyplot as plt)
* Seaborn (import seaborn as sns)

Do not write any import statements (or include comments regarding imports)
Do not try to simulate any data
All files must be written to /tmp/ directory
Do not plt.show() the plot, just save it
Do not add any return variables
Do not make checks to validate presense of columns, you already have that info below
If there are plots, unless explicitly asked use vertical formats. Plot labels could be long, so add
enough padding with `bbox_inches='tight'` when saving figures
If the chosen fields are not strings apply following rules
- dict - a dictionary with keys "id" and "label" mark this in feedback to let usser adjust code
- list - list entries likely are dictionaries with keys "id" and "label" mark this in feedback to
let user adjust code

INPUT
Following inputs are available for you. All inputs are pandas dataframes
{data}

QUERY
{query}
```

The expected output from the LLM is a JSON with 'code', 'files', 'assumptions' and 'feedback' attributes. Files are used to capture the plots, CSVs and TSVs created by the code execution. These attributes are used to generate the final python script. Note that, non-Pythonic information is appended to the script file as comment blocks.

The data field inject the format of the Pandas data frame as 'str', 'list' or 'dict' depending on the datatype of the columns in the data frame and hints the LLM to act accordingly. The query will carry the original analysis description by the user.